\documentclass{article}

\usepackage[preprint]{neurips_2025}
\usepackage{enumitem}
\usepackage{soul}
\setlist[itemize]{nosep}
\setlist[enumerate]{nosep}
\usepackage{xcolor}
\usepackage{hyperref}

\usepackage{graphicx} 
\usepackage{subcaption}
\usepackage{wrapfig}
\usepackage{graphicx}
\usepackage{caption}
\usepackage{booktabs}
\usepackage{multirow}
\usepackage{algorithm}
\usepackage{algpseudocode}
\usepackage{amsmath}

\usepackage[most]{tcolorbox}
\usepackage{xcolor}  
\newtcolorbox{prompt}[1]{
  enhanced,
  colback=blue!5!gray!10,
  colframe=blue!50!black,
  arc=2mm,
  boxrule=1pt,
  title=#1,
  fonttitle=\bfseries\color{white},
  coltitle=green!50!black,
  breakable=false,
  width=\textwidth
}

\usepackage{geometry}
\usepackage{inconsolata}
\newtcolorbox{promptbox}[1][]{
  colback=gray!5,
  colframe=black!50,
  boxrule=0.8pt,
  arc=2mm,
  fontupper={\ttfamily\small},
  enhanced,
  breakable,
  left=5pt,
  right=5pt,
  top=5pt,
  bottom=5pt,
  title=#1
}
\newcommand{\concat}{\,\|\,} 
\newcommand{\Embd}{\mathsf{Embd}}
\algnewcommand{\SectionComment}[1]{\Statex \textit{// #1}}

\definecolor{color1}{rgb}{0.1,0.7,0.8} 
\definecolor{color2}{rgb}{0.9,0.1,0.1} 
\definecolor{color3}{rgb}{0.7,0.3,0.7} 
\definecolor{color4}{rgb}{0.3,0.3,0.7} 
\definecolor{color5}{RGB}{8, 102, 3} 
\definecolor{color6}{rgb}{0.53, 0.66, 0.42} 

\usepackage[utf8]{inputenc} 
\usepackage[T1]{fontenc}    
\usepackage{hyperref}       
\usepackage{url}            
\usepackage{booktabs}       
\usepackage{amsfonts}       
\usepackage{nicefrac}       
\usepackage{microtype}      
\usepackage{xcolor}         

\title{CoIn: \underline{Co}unting the \underline{In}visible Reasoning Tokens in Commercial Opaque LLM APIs}

\author{
\textbf{Guoheng Sun\textsuperscript{1}}\thanks{ghsun@umd.edu}\hspace{0.2em}, 
\textbf{Ziyao Wang\textsuperscript{1}}, 
\textbf{Bowei Tian\textsuperscript{1}}, \\
\textbf{Meng Liu\textsuperscript{1}},
\textbf{Zheyu Shen\textsuperscript{1}}, 
\textbf{Shwai He\textsuperscript{1}}, 
\textbf{Yexiao He\textsuperscript{1}}, \\ 
\textbf{Wanghao Ye\textsuperscript{1}},
\textbf{Yiting Wang\textsuperscript{1}}, 
\textbf{Ang Li\textsuperscript{1}}\thanks{angliece@umd.edu}  \\
\textsuperscript{1}University of Maryland, College Park
}

\newcommand{\n}{\texttt{CoIn}}

\begin{document}

\maketitle

\begin{abstract}
As post-training techniques evolve, large language models (LLMs) are increasingly augmented with structured multi-step reasoning abilities, often optimized through reinforcement learning. These reasoning-enhanced models outperform standard LLMs on complex tasks and  now underpin many commercial LLM APIs. However, to protect proprietary behavior and reduce verbosity, providers typically conceal the reasoning traces while returning only the final answer. This opacity introduces a critical transparency gap: users are billed for invisible reasoning tokens, which often account for the majority of the cost, yet have no means to verify their authenticity. This opens the door to \emph{token count inflation}, where providers may overreport token usage or inject synthetic, low-effort tokens to inflate charges. To address this issue, we propose \texttt{CoIn}, a verification framework that audits both the \textit{quantity} and \textit{semantic validity} of hidden tokens. \texttt{CoIn} constructs a verifiable hash tree from token embedding fingerprints to check token counts, and uses embedding-based relevance matching to detect fabricated reasoning content. Experiments demonstrate that \texttt{CoIn}, when deployed as a trusted third-party auditor, can effectively detect token count inflation with a success rate reaching up to 94.7\%, showing the strong ability to restore billing transparency in opaque LLM services. 
The dataset and code are available at \url{https://github.com/CASE-Lab-UMD/LLM-Auditing-CoIn}.

\end{abstract}

\section{Introduction}

Large language models (LLMs) have achieved significant advances in recent years. Yet, as pretraining begins to saturate available data resources~\cite{zoph2020rethinking}, the research community has increasingly turned to inference-time innovations~\cite{hu2023llm,kumar2025llm}. Among these, reinforcement learning (RL)-optimized reasoning models have shown promise by generating longer, structured reasoning traces that improve performance, particularly in tasks involving mathematics and code~\cite{guo2025deepseek,muennighoff2025s1}. Such models, exemplified by DeepSeek-R1~\cite{guo2025deepseek} and ChatGPT-O1~\cite{jaech2024openai}, demonstrate that scaling at inference time can yield new capabilities without further pretraining.

With this shift, providers like OpenAI increasingly adopt new service models. Reasoning traces, while critical for quality, are often verbose, sometimes speculative~\cite{jin2024impact,zhang2025lightthinker}, and may reveal internal behaviors vulnerable to distillation~\cite{gou2021knowledge,sreenivas2024llm}. To protect proprietary methods and streamline outputs, commercial APIs typically suppress these intermediate steps, exposing only the final answer. However, users are still charged for all generated tokens, including those hidden from view. We refer to such services as \textbf{Commercial Opaque LLM APIs (COLA)}—proprietary, pay-per-token APIs that conceal both intermediate text and logits.

\begin{figure}[t]
\centering
\begin{subfigure}{0.44\textwidth}
    \centering
    \includegraphics[width=6.2cm,height=3.9cm,keepaspectratio=false]{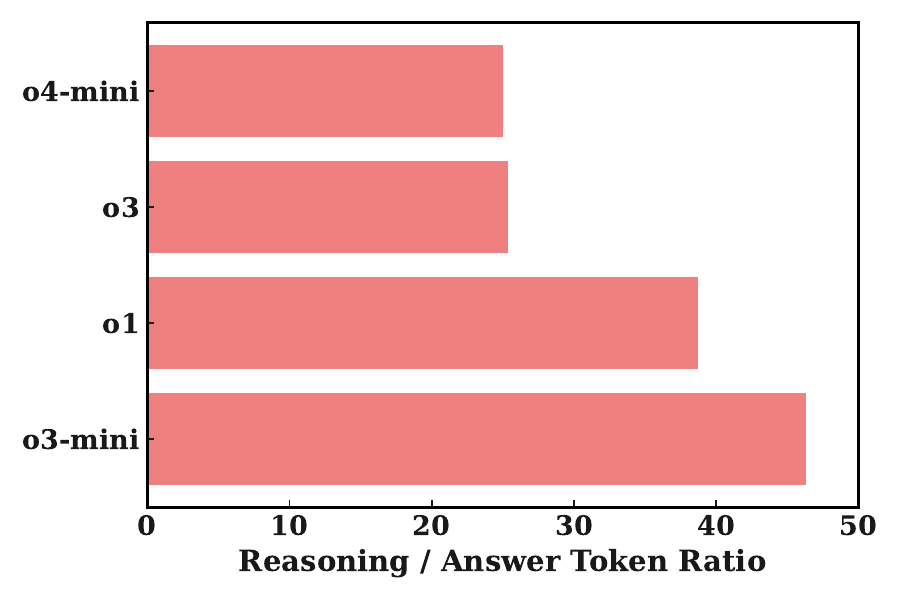}
    \caption{}
    \label{fig:token_ratio_datasets}
\end{subfigure}
\begin{subfigure}{0.44\textwidth}
    \centering
    \includegraphics[width=6.9cm,height=3.9cm,keepaspectratio=false]{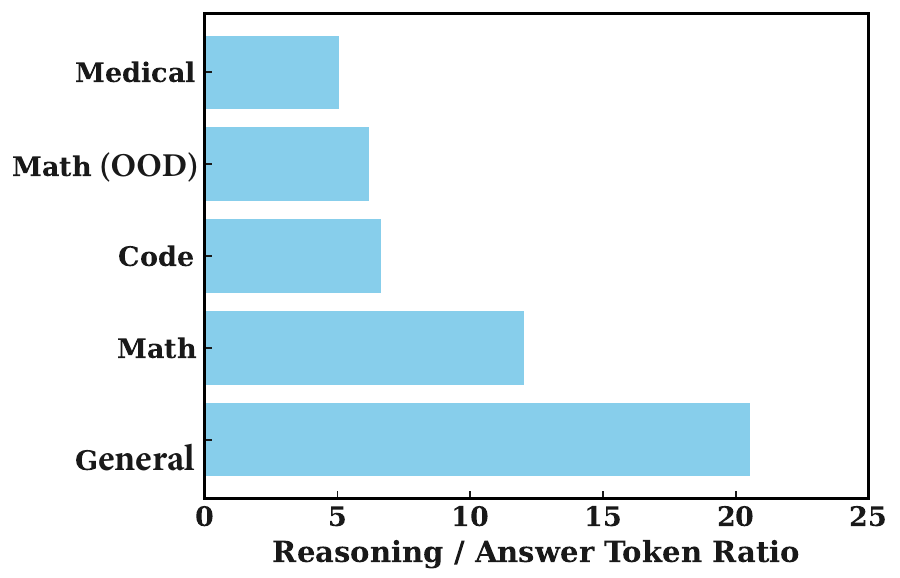}
    \caption{}
    \label{fig:token_ratio_apis}
\end{subfigure}
\caption{Ratio of reasoning tokens to answer tokens across datasets and deployed APIs.
(a) Token ratios on the OpenR1-Math dataset across different OpenAI reasoning models.
(b) Token ratios of the DeepSeek-R1~\cite{deepseekai2025deepseekr1incentivizingreasoningcapability} across various reasoning datasets.
In both cases, the number of reasoning tokens often exceeds answer tokens by an order of magnitude or more.}
\label{fig:token_ratio_comparison}
\end{figure}

This design introduces a critical vulnerability: users have no means to verify token usage or detect overbilling. Because reasoning tokens often outnumber answer tokens by more than an order of magnitude (Figure~\ref{fig:token_ratio_comparison}), this invisibility allows providers to \textbf{misreport token counts} or \textbf{inject low-cost, fabricated reasoning tokens to artificially inflate token counts}. We refer to this practice as \textbf{token count inflation}. For instance, a single high-efficiency ARC-AGI run by OpenAI's o3 model consumed 111 million tokens, costing \$66,772.\footnote{\url{https://arcprize.org/blog/oai-o3-pub-breakthrough}} Given this scale, even small manipulations can lead to substantial financial impact. Such information asymmetry allows AI companies to significantly overcharge users, thereby undermining their interests.

To tackle this problem, we design \textbf{{\n} (Counting the Invisible)}, a verification framework  that enables third-party auditing of invisible reasoning tokens in \textsc{COLA} services. {\n} ensures billing accountability by enabling users to validate token counts reported by the commercial provider, while preserving the confidentiality of hidden content and maintaining protection against distillation. 

{\n} consists of two key components:
\textbf{(1) Token Quantity Verification}, which leverages a verifiable hash tree~\cite{merkle1987digital} to store fingerprint embeddings of reasoning tokens. Upon an audit request, {\n} allow users to query a small subset of the token fingerprints in the hash tree to verify the number of invisible tokens, avoiding accessing the actual reasoning tokens; and
\textbf{(2) Semantic Validity Verification}, which detects fabricated, irrelevant, or low-effort token injection via a semantic relevance matching head. This matching head takes the embeddings of both the reasoning tokens and the answer tokens as input, and outputs a relevance score indicating their semantic consistency. Users can assess this score to identify token count inflation with low-effort token injection.
Together, these components enable {\n} to identify misreported token counts and fabricated reasoning traces, enabling transparent billing without exposing proprietary data.
In practice, {\n} can be deployed as a trusted third-party auditing service that ensures billing transparency while preserving the integrity and confidentiality requirements of COLA providers.

Our main contributions are as follows:
\begin{itemize}
    \item We define the COLA architecture and formalize the emerging threat of \emph{token count inflation}, categorizing both naive and adaptive inflation strategies.
    \item We design {\n}, a verification framework combining \emph{token quantity verification} via verifiable hashing and \emph{semantic validity verification} via embedding relevance, to audit invisible reasoning tokens without exposing proprietary content.
    \item Our experiments demonstrate that {\n} can achieve a 94.7\% detection success rate against various adaptive attacks with less than 40\% embedding exposure and less than 4\% token visibility. Moreover, even when 10\% of tokens are maliciously forged by COLA, {\n} still maintains a 40.1\% probability of successful detection.

\end{itemize}

\section{Related Work}
\noindent \textbf{Reasoning Model.}
LLMs have shown strong performance on complex reasoning tasks by generating intermediate steps, a technique known as chain-of-thought prompting~\cite{wei2022chain}. This paradigm has been further enhanced by methods such as self-consistency~\cite{wang2022self} and program-aided reasoning~\cite{gao2023pal}. Recent research reveals that generating more reasoning steps at inference time can lead to higher answer accuracy, a phenomenon referred to as the test-time scaling law~\cite{snell2024scaling}, which has become a guiding principle for optimizing LLMs. Reasoning models are typically LLMs fine-tuned via RL~\cite{rafailov2023direct,wu2023pairwise,ramesh2024group} to produce structured reasoning traces before generating final answers, thereby improving answer quality. These reasoning traces are often longer, more indirect, and may include failed attempts, but are nonetheless closely tied to the final answer~\cite{hao2024llm,yang2025understanding}. Since these reasoning tokens are generated in the same autoregressive manner as answer tokens, COLA charge for them based on token count. However, the indirect and verbose nature of reasoning makes it challenging to audit their legitimacy without direct access to the reasoning traces themselves.

\noindent \textbf{COLA Auditing.}
Several works have emerged to address the lack of transparency in COLA. \cite{cai2025you} proposes a watermark-based method to audit whether a COLA uses the required LLM rather than a cheaper LLM. Similarly, \cite{yuan2025trust} develops a user-verifiable protocol to detect nodes that run unauthorized or incorrect LLM in a multi-agent system. Another series of works~\cite{zheng2025calm,marks2025auditing} proposes auditing some bad behaviors of LLMs, e.g., cheating and offensive outputs. These techniques mainly focus on the model auditing and lack attention to the token count auditing of COLA.

\section{Preliminary}

\noindent \textbf{Participants and Problem Formulation.}
The {\n} framework involves three roles: (i) COLA — a commercial LLM service provider (e.g., OpenAI) that performs multi-step reasoning and returns only the final output to the user;
(ii) User — an end-user who submits a prompt and receives an answer along with a billing summary;
and (iii) {\n} auditor — a trusted third party responsible for verifying the invisible reasoning tokens on behalf of the user.

In each service interaction, the user sends a prompt $P$ to COLA. The LLM generates reasoning tokens $R = \{r_1, r_2, \dots, r_m\}$, followed by answer tokens $A = \{a_1, a_2, \dots, a_n\}$.
Only the final answer $A$ is returned to the user, while the reasoning trace $R$ remains hidden. Billing is based on the total number of tokens $m + n$, including the invisible reasoning tokens. As Figure~\ref{fig:token_ratio_comparison} shows, reasoning tokens often dominate the total count, i.e., $m \gg n$, resulting in a significant transparency gap.

\noindent  \textbf{Token Count Inflation.}
We consider two strategies for inflating token counts:

\begin{itemize}[leftmargin=1em]
    \item \textbf{Naive token count inflation.} COLA reports a falsified token count 
    $m_f > m$, leading to direct overbilling without modifying the output.

    \item \textbf{Adaptive token count inflation.} Anticipating user-side defenses (e.g., hash matching, spot-checking), COLA may append low-effort fabricated reasoning tokens to the original reasoning trace. These fabricated tokens can be generated via random sampling, retrieval from related documents, or repetition of existing tokens, and then indistinguishably mixed with genuine reasoning tokens. The inflated sequence is then used for billing, bypassing naive verification methods and still overcharging the user.
\end{itemize}

To address these threats, {\n} employs two components: (1) \textbf{Token Quantity Verification}, which audits the reported token count using verifiable commitments and exposes embeddings; and 
(2) \textbf{Semantic Validity Verification}, which evaluates the relevance between 
reasoning and answer tokens to detect low-quality injections.

\noindent  \textbf{Threat Model.}
\textsc{COLA} has access to the user prompt $P$, the full reasoning trace $R$, 
and the answer $A$, and controls the billing report $(m, n)$, where $m$ is the claimed number of reasoning tokens and $n$ is the number of answer tokens. It can manipulate the reported count without user visibility.
The {\n} auditor operates as a trusted third party. It can access $P$, $A$, and 
$(m, n)$. but cannot observe $R$ directly or directly query the LLM used by COLA. However, it can request COLA to return the embeddings of $R$, computed using an embedding 
model fixed by the auditor to prevent tampering.

\section{{\n}: Counting the Invisible Reasoning Tokens}

\begin{figure}[tbp]
    \centering
    \includegraphics[width=1.02\textwidth]{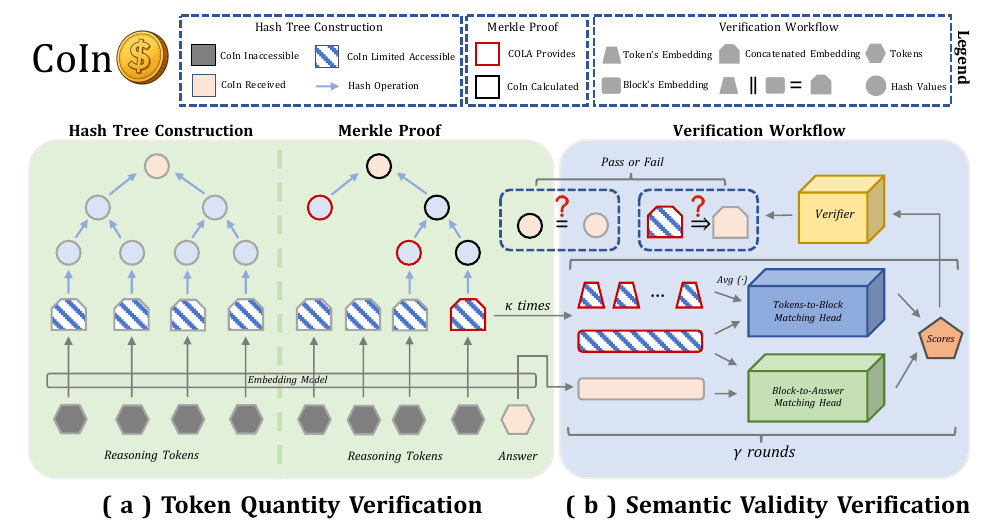} 
    \caption{{\n} Framework. 
    }
    \label{fig:coin_framework}
\end{figure}

{\n} comprises two complimentary components: \textbf{token quantity verification} and 
\textbf{semantic validity verification}.
The token quantity verification module treats embeddings of invisible reasoning 
tokens as cryptographic fingerprints and organizes them into a verifiable hash tree. By querying a small subset of these fingerprints, users can audit the claimed number of invisible tokens without accessing their contents.
The semantic validity verification module trains a lightweight neural network, referred to as a \emph{matching head}, to evaluate the relevance between embeddings. During auditing, {\n} retrieves token embeddings from the hash tree and uses the matching head to compute relevance scores both among reasoning tokens and between reasoning  and answer tokens. These scores help detect token count inflation through the injection of fabricated or irrelevant reasoning tokens.
An overview of the {\n} framework is illustrated in Figure~\ref{fig:coin_framework}.

\subsection{Token Quantity Verification}

\noindent \textbf{Token Fingerprint Generation.}
In {\n}, COLA is required to generate embeddings of its reasoning tokens using a third-party embedding model $\mathsf{Embd}(\cdot)$ designated by the {\n} auditor. These embeddings serve as token fingerprints used to construct 
a verifiable hash tree for auditing. This hash tree enables {\n} to audit the total number of invisible tokens without accessing the tokens themselves.

Specifically, given a reasoning sequence $R$, COLA first partitions $R$ into $\alpha$ 
blocks. For each token $r_i$ in block $B_j$, COLA computes:  
(i) the block embedding $\mathsf{Embd}(B_j)$, which embeds all the tokens inside the block; and 
(ii) the token embedding $\mathsf{Embd}(r_i)$, which embeds the single token itself.  
Each reasoning token therefore acquires both the block embedding and the token embedding. For each reasoning token $\mathsf{Embd}(r_i)$, {\n} concatenated its block embedding and token embedding to form the token fingerprint:
$\mathsf{Embd}(B_j) \,\|\, \mathsf{Embd}(t_i)$.

\noindent  \textbf{Fingerprint Hash Tree Construction.} COLA applies a cryptographic hash function (e.g., SHA-256), agreed upon with {\n}, to each token fingerprint to construct the leaf nodes of a 
Merkle Hash Tree~\cite{merkle1987digital}.
The number of leaf nodes is padded to the nearest power of two, and parent nodes are built recursively 
by hashing concatenated sibling nodes up to the Merkle Root. This root serves as a commitment to the full set of reasoning tokens and is submitted 
to {\n}.
After constructing the hash tree, COLA give the Merkle Root to {\n} for Merkle Proofs upon user's auditing request.

\noindent  \textbf{Merkle Proof.}
Upon receiving the answer $A$ and the token counts $m$ and $n$, a user may suspect token inflation. To verify the count of invisible reasoning tokens, 
the user selects a block $B_j$ and randomly chooses token indices to audit.  
Upon receiving the request, {\n} auditor requests the following information from COLA:
(i) the fingerprints of the selected tokens; and  
(ii) the corresponding Merkle Path, which is a sequence of sibling hashes needed to reconstruct the Merkle Root from the corresponding token.
{\n} recomputes the Merkle root from the provided data and checks for consistency with the original commitment by COLA provider. A successful match confirms the integrity of the selected token; a mismatch indicates possible fabrication and inflated token reporting.
The construction and Merkle Proof procedure is illustrated in Figure~\ref{fig:coin_framework}-(a) and detailed further in Appendix~\ref{app:tree} , \ref{app:proof}.

The Merkle proof in token quantity verification ensures both the structural integrity and the correctness of the reported token count, effectively defending against naive token count inflation. However, a dishonest COLA may still conduct adaptive token count inflation by injecting irrelevant or low-effort fabricated tokens that pass count verification. To address this limitation, we introduce semantic validity verification.

\subsection{Semantic Validity Verification}
\label{sec:semantic-validity}
To defend against adaptive token count inflation, we introduce the semantic validity verification component, as illustrated in Figure~\ref{fig:coin_framework}-(b).  
This component ensures that reasoning tokens are semantically meaningful and contribute to the final answer, preventing low-effort or fabricated 
token insertion.
Based on this principle, {\n} verifies the semantic validity of invisible tokens from two perspectives:  
\begin{itemize}[leftmargin=1em]
    \item \textbf{Token-to-Block verification} checks whether each reasoning token $r_i$ is semantically coherent within its enclosing block $B_j$. This defends against randomly injected or meaningless tokens.
    \item \textbf{Block-to-Answer verification} evaluates whether a reasoning block $B_j$ is semantically aligned with the final answer $A$, thus identifying the insertion of low-cost content that is insufficiently relevant to the task.
\end{itemize}

To support both tasks, {\n} trains two lightweight neural modules called the \textit{matching heads}, which are binary classifiers that determines whether two embeddings are semantically associated.  
Given two token embeddings $a$ and $b$, the matching head first computes the cosine similarity:
\(
\text{cos\_sim} = \frac{a \cdot b}{\|a\| \|b\|},
\)
and constructs the feature vector:
\[
h = [a;\, b;\, a - b;\, a \odot b;\, \text{cos\_sim}] \in \mathbb{R}^{4d + 1},
\]
where $d$ is the embedding dimension, $[\,;\,]$ denotes concatenation, and $\odot$ denotes element-wise multiplication. The feature $h$ is then passed through a two-layer feedforward network to produce a scalar match score $S \in [0,1]$, representing the likelihood that $a$ and $b$ are semantically aligned. This process can be viewed as a regression function $S = \mathsf{MH}(a, b)$.

In {\n}, the matching heads $\mathsf{MH_{tb}}(\cdot), \mathsf{MH_{ba}}(\cdot)$ are trained offline for token-to-block and block-to-answer verification respectively. {\n} use open-source corpora and the same embedding model in token fingerprinting to build the datasets for matching heads training.

\noindent  \textbf{Verification Protocol.}
In each verification round, the user randomly selects some reasoning tokens $r_i$ (by default, 10\% of the tokens within a selected block) from the hash tree. 
Since the token fingerprint consists of both the token embedding $\mathsf{Embd}(r_i)$ and the corresponded block embedding $\mathsf{Embd}(B_j)$, it can be directly used for Tokens-to-Block verification. For the Block-to-Answer verification, we use $\mathsf{Embd}(B_j)$ and the embedding of the whole answer to compute the score:
\begin{equation}
    S_{tb} = \mathsf{MH_{tb}}(\mathsf{AVG}(\mathsf{Embd}(r_i)), \mathsf{Embd}(B_j)), \qquad
    S_{ba} = \mathsf{MH_{ba}}(\mathsf{Embd}(B_j), \mathsf{Embd}(A)).
\end{equation}

Here, $S_{tb}$ and $S_{ba}$ represent the relevance scores for the two respective verification tasks.
Each score reflects the estimated likelihood that the two input embeddings are semantically relevant.

\subsection{Workflow of {\n}}

\noindent  \textbf{Enforcing Billing Integrity with {\n}.}
When a user suspects token count inflation in a specific response, they can initiate an audit request to {\n}.
The audit begins with the user selecting 
a fraction $\gamma$ of the total reasoning blocks for verification.
{\n} then performs two Semantic Validity Verifications and multiple Merkle Proofs on these selected blocks.
The resulting match scores are passed to a verifier, which issues a final decision. If the verifier accepts, the audit concludes successfully.
If the verifier rejects, the user continues by randomly selecting another unverified block for auditing. This process repeats until either a successful judgment is reached or all blocks are exhausted. If no verification passes, the audit concludes with COLA being flagged for token inflation. The user may then request COLA to justify the charges by disclosing the original 
reasoning content.
The complete procedure is outlined in Algorithm~\ref{alg:multi_round_verification}.

\noindent \textbf{Verifier Design.} Each audit round produces a variable-length sequence of match scores, as the number of verified blocks depends on verifier decisions. To handle this, we implement two types of verifiers:
(i) \textbf{Rule-based}: Averages the scores from two semantic verifications. The audit passes if both averages exceed a threshold $\tau$.
(ii) \textbf{Learning-based}: Uses a lightweight DeepSets model~\cite{zaheer2017deep} to process the unordered set of match scores and audit will succeed if the confidence exceeds $\tau$.

Auditing outcomes enable users to assess the trustworthiness of a COLA provider. 
Frequent failures in {\n} audits may erode user trust and damage provider reputation. By introducing verifiable accountability, the {\n} framework serves as a deterrent against token count inflation in commercial LLM services.

\noindent \textbf{Hyperparameter and Verification Cost.} 
{\n} is governed by a few hyperparameters that control auditing granularity and cost. Specifically, $\alpha$ is the number of blocks, $\beta$ the block size, $\gamma$ the initial sampling ratio (default: 0.3), and $k$ the number of tokens sampled per block (default: $0.1 * \beta$). A smaller $\beta$ reduce exposure but increase overhead. The protocol begins with $\gamma \cdot \alpha$ rounds and may proceed up to $\alpha$ rounds under early stopping, so the number of verification rounds satisfies 
\(
\ell \in [\gamma \cdot \alpha,\ \alpha]
\).
As a result, the total number of Merkle Proofs is 
\(
k \cdot \ell
\),  
and the number of Semantic Judgments is 
\(
2 \cdot \ell
\).

\section{Experiments}
\label{sec:experiments}
We systematically evaluate the robustness and reliability of {\n} and its submodules under various adaptive inflation attacks across multiple datasets. We further analyze the construction cost of the Hash Tree, as well as whether the partially exposed block embeddings and tokens can be exploited to recover the reasoning tokens of COLA. Finally, we assess the difficulty of the dataset we constructed.

\subsection{Experiment Setup}

\noindent \textbf{Token Inflation Implementations.}  
We study both \textit{naive} and \textit{adaptive} token count inflation strategies. To enable fine-grained evaluation and systematic dataset construction, we design four variants of adaptive inflation. All inflation types used in our experiments are summarized in Table~\ref{tab:inf_exp}. These strategies are applied to generate inflated samples for both training and evaluation.
\vspace{-0.2in}
\begin{table}[h]
\centering
\caption{Token inflation types used in our experiments.}
\begin{tabular}{ll}
\toprule
\textbf{Type} & \textbf{Description} \\
\midrule
\textbf{Naive Inflation} & Randomly select tokens from the vocabulary for injection. \\
\textbf{Ada. Inflation 1} & Inject tokens with embeddings similar to $P$, $R$, or $A$. \\
\textbf{Ada. Inflation 2} & Inject tokens directly sampled from $P$, $R$, or $A$. \\
\textbf{Ada. Inflation 3} & Inject reasoning sequences extracted from other inputs. \\
\textbf{Ada. Inflation 4} & Inject retrieved sequences semantically similar to $P$, $R$, or $A$. \\
\bottomrule
\end{tabular}

\label{tab:inf_exp}
\end{table}

\noindent \textbf{Datasets and Training Setup.}  
We conduct experiments on five datasets derived from \texttt{DeepSeek-R1}~\cite{deepseekai2025deepseekr1incentivizingreasoningcapability}, covering diverse reasoning domains: medical~\cite{chen2024huatuogpto1medicalcomplexreasoning}, code~\cite{OpenThoughts,openr1}, mathematics~\cite{openr1}, general reasoning\footnote{\url{https://huggingface.co/datasets/glaiveai/reasoning-v1-20m}}, and out-of-domain (OOD) mathematics~\cite{OpenThoughts,openr1}. For training, we randomly sample 20,000 examples from each dataset and combine them into a joint dataset. Another 1,000 samples per dataset are held out to form the evaluation set for the {\n} framework. We use the tokenizer of \texttt{DeepSeek-R1} in our experiments.

For the matching head, we use \texttt{all-MiniLM-L6-v2}~\cite{reimers-2019-sentence-bert} as model structure. In the \textbf{token-to-block} verification task, we treat original samples as normal instances and apply Naive Inflation as well as Adaptive Inflation 1 and 2 to construct inflated samples. Normal and inflated samples are labeled 0 and 1 respectively and mixed at a 1:1 ratio to form the training set. For the \textbf{block-to-answer} verification task, we adopt a similar setup, but use Ada. Inflation 1--4 to construct inflated samples. This setting enables a thorough evaluation of the model's ability to detect both shallow and semantically sophisticated inflation attacks. The details of training and dataset are explained in Appendix~\ref{apdx:data}, \ref{apdx:train}.

\noindent \textbf{Metrics.}
We define the \textit{Detection Success Rate (DSR)} as the classification accuracy of our module, computed separately for malicious samples and benign samples. 
\textit{Inflation Rate (IR)} indicates the percentage of fabricated tokens injected by COLA relative to the number of original reasoning tokens. For benign samples, the \textit{Average Exposure Rate (AER)} refers to the proportion of blocks exposed during the {\n} multi-step verification process out of the total number of blocks.

\begin{figure}[tbp]
    \centering
    \includegraphics[width=1.07\textwidth]{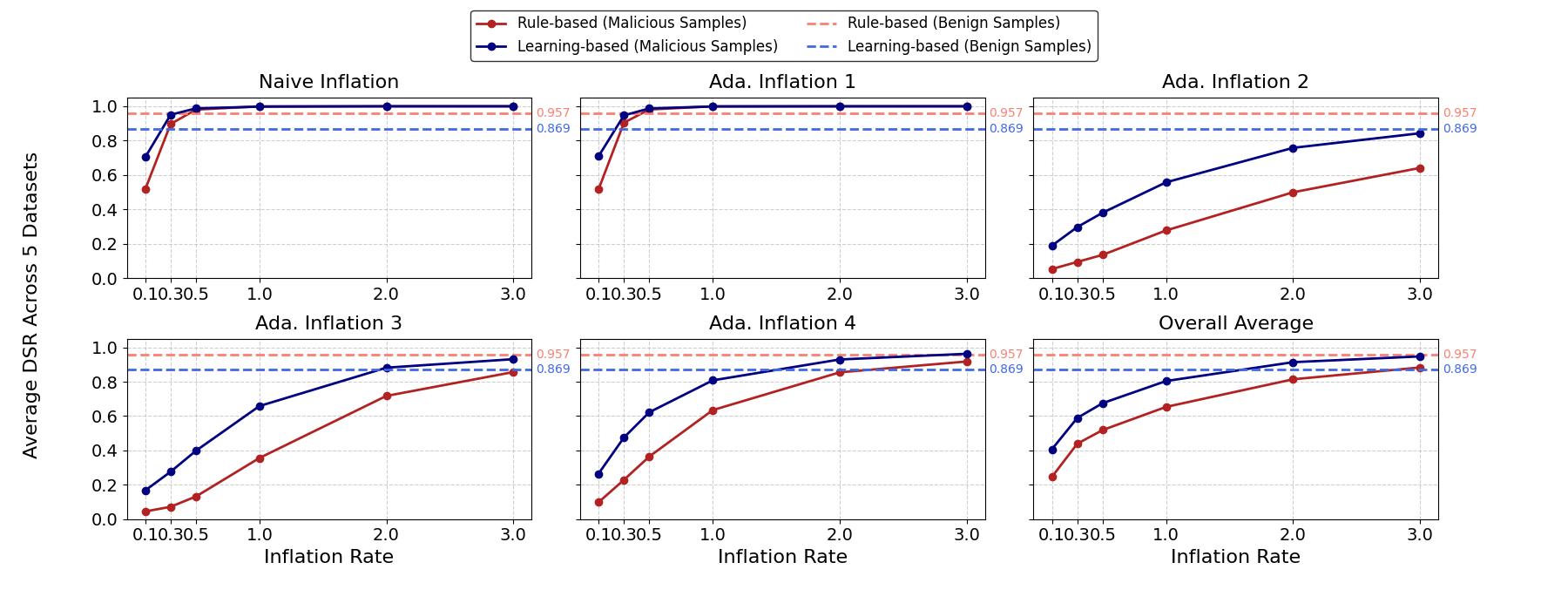} 
    \caption{Performance of {\n} across different inflation methods and verifiers. The red lines and the blue lines represent the DSR of rule-based verifier and learning-based verifier, respectively. $\gamma$
    }
    \label{fig:main_coin1}
\end{figure}

\subsection{Detection Performance of {\n}}
\begin{figure}[tbp]
    \centering
    \includegraphics[width=1.05\textwidth]{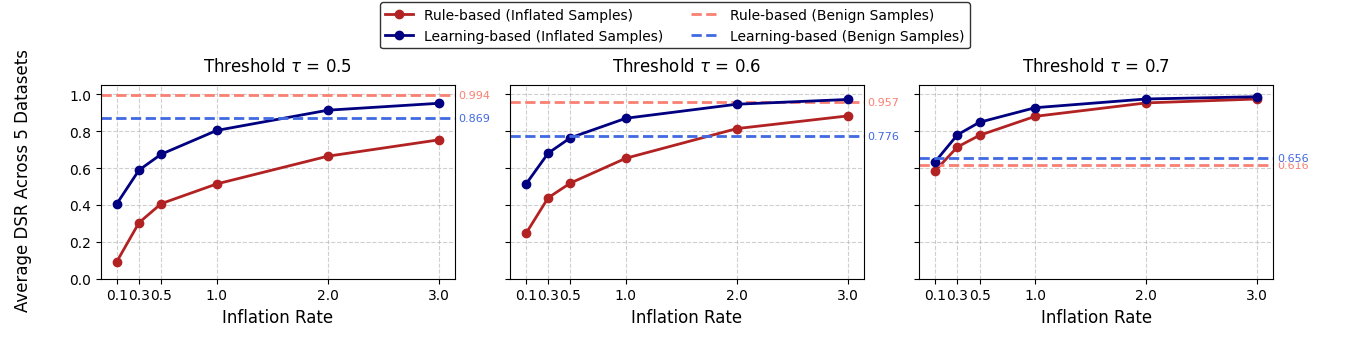} 
    \caption{Impact of threshold $\tau$ on DSR. 
    }
    \label{fig:main_coin2}
\end{figure}

We evaluate {\n}'s ability to detect various token count inflation attacks. Figure~\ref{fig:main_coin1} shows the relationship between IR and DSR across five datasets. DSR increases with IR, indicating that more aggressive inflation is easier to detect. For example, in the Naive Inflation setting, DSR 
approaches 100\% when IR exceeds 0.5, suggesting near-perfect detection for heavily inflated queries. 
Figure~\ref{fig:main_coin2} illustrates the impact of $\tau_h$. As $\tau_h$ increases, DSR for malicious samples increases, while for benign samples decreases. This highlights a tunable trade-off between user utility and COLA's protection.

\noindent \textbf{Learning-based Verifier Excels at Detecting Malicious Samples.} For a fair comparison, we set the threshold $\tau = 0.5$ for learning-based verifier, and $\tau = 0.6$ for rule-based verifier. The initial sampling ratio $\gamma$ is set to 0.3. Across all settings, the learning-based verifier consistently outperforms the rule-based variant in detecting inflated samples, achieving up to 94.7\% average DSR at an inflation ratio (IR) of 3.0. Among the adaptive methods, Ada. Inflation 2 and Ada. Inflation 3 present greater detection challenges. Nevertheless, {\n} still achieves an average DSR of approximately 84.3\% and 93.1\% at IR = 3.0 for these cases. These results demonstrate that {\n} remains robust even under strong adaptive inflation, and that learning-based semantic verification offers significant advantages in practical detection scenarios.

\begin{wraptable}{r}{0.35\linewidth}
    \centering
    \vspace{-10pt}
    \caption{Influence of Block Size.}
    \label{tab:evaluation_AER}
    \begin{tabular}{lccc}
        \toprule
        \multirow{2}{*}{Metric} & \multicolumn{3}{c}{Block Size $\beta$} \\
        \cmidrule(lr){2-4}
                                & 256 & 512 & 1024 \\
        \midrule
        Avg. $\alpha$                & 16.8 & 8.6 & 4.5 \\
        Avg. $\ell$     & 6.3 & 3.7 & 2.2 \\
        AER~$\downarrow$        & 0.38 & 0.43 & 0.49 \\
        \bottomrule
    \end{tabular}
    \vspace{-10pt}
\end{wraptable}

\noindent \textbf{Rule-based Verifier Excels at Handling Benign Samples.} Although it is less effective at detecting inflated samples, the rule-based verifier performs better on benign samples when properly tuned, due to its lower false positive rate. Given our verification mechanism, a higher DSR on benign samples results in a lower Average Exposure Rate (AER), which makes COLA more favorable toward such mechanisms, potentially even negotiating certain parameters with users. Based on the rule-based verifier ($\tau = 0.6$), Table~\ref{tab:evaluation_AER} reports the average number of blocks $\alpha$, average verification rounds $\ell$ and AER across different block sizes $\beta$. With increasing $\beta$, $\ell$ decreases, indicating lower cost. However, AER increases, and the results in Appendix~\ref{apdx:blocks} show that increasing the block size reduces the DSR for adaptive inflation.

\subsection{Performance of the Semantic Validity Verification}\label{sec:SVV}

\noindent \textbf{Block-to-Answer Verification Task.} We separately evaluate the performance of the two types of matching heads introduced in Section~\ref{sec:semantic-validity}. Table~\ref{tab:B2A_main} shows the DSR of the matching heads on the \textit{Block-to-Answer Verification} task. 
The model achieves an average DSR of 94.8\% across attack types. Even for the Math (OOD) dataset, 
which was excluded from training, the model performs strongly, indicating good generalization.
The DSR drops slightly on clean (non-inflated) samples due to the presence of reasoning blocks not directly contributing to the final answer (see Section~\ref{sec:dis-hard-data}). Additionally, Content Reuse 2 attacks introduce hard negatives that resemble real data, making separation more difficult.

\begin{table*}[htbp]
\renewcommand{\arraystretch}{1.0} 
\centering
\caption{Block-to-Answer Verification Performance Across Attack Types and Domains.}
\resizebox{0.85\linewidth}{!}{
\begin{tabular}{l|ccccc|c}
    \toprule
    \bf Attack Type & \bf Medical & \bf Code & \bf Math & \bf General & \bf Math (OOD) & \bf Avg. \\
    \midrule
    \textbf{Naive Inflation}   & 99.4 & 100.0 & 100.0 & 99.3 & 100.0 & 99.7 \\
    \textbf{Ada. Inflation 1}   & 95.3 & 98.7 & 98.6 & 96.8 & 98.2 & 97.5 \\
    \textbf{Ada. Inflation 2}      & 94.4 & 92.3 & 92.8 & 94.2 & 92.7 & 93.3 \\
    \textbf{Ada. Inflation 3}      & 89.2 & 81.5 & 84.3 & 92.9 & 84.6 & 86.5 \\
    \textbf{Ada. Inflation 4}    & 94.2 & 97.9 & 99.0 & 96.1 & 97.8 & 97.0 \\
    \midrule
    \bf Avg. With Inflation  & 94.5 & 94.1 & 94.9 & 95.8 & 94.7 & 94.8 \\
    \midrule
    \bf No Inflation    & 87.9 & 90.3 & 87.1 & 86.5 & 87.9 & 87.9 \\
    \midrule
\end{tabular}
}
\label{tab:B2A_main}
\end{table*}

\noindent \textbf{Tokens-to-Block Verification Task.}
Table~\ref{tab:attack_domain_results_trimmed} shows the results for token-to-block verification. The model performs well overall but struggles with Adaptive Inflation 2, where tokens reused from the same sample lead to significant lexical and semantic overlap. This overlap can blur the distinction between original and fabricated content, especially when reused tokens legitimately contribute to the block.

\begin{table*}[htbp]
\renewcommand{\arraystretch}{1.0} 
\centering
\caption{Tokens-to-Block Verification Performance Across Attack Types and Domains.}
\resizebox{0.85\linewidth}{!}{
\begin{tabular}{l|ccccc|c}
    \toprule
    \bf Attack Type & \bf Medical & \bf Code & \bf Math & \bf General & \bf Math (OOD) & \bf Avg. \\
    \midrule
    \textbf{Naive Inflation} & 90.8 & 90.5 & 95.3 & 84.5 & 94.6 & 91.2 \\
    \textbf{Ada. Inflation 1} & 95.1 & 96.1 & 95.8 & 95.5 & 95.8 & 95.6 \\
    \textbf{Ada. Inflation 2} & 76.0 & 75.2 & 73.9 & 73.6 & 74.8 & 74.7 \\
    \midrule
    \bf Avg. With Inflation & 87.3 & 87.2 & 88.4 & 84.5 & 88.4 & 87.2 \\
    \midrule
    \bf No Inflation & 82.0 & 80.4 & 87.2 & 79.0 & 86.0 & 82.9 \\
    \bottomrule
\end{tabular}
}
\label{tab:attack_domain_results_trimmed}
\end{table*}

\begin{figure}[ht]
    \centering
    \begin{minipage}[c]{0.48\linewidth}
        \centering
        \includegraphics[width=\linewidth]{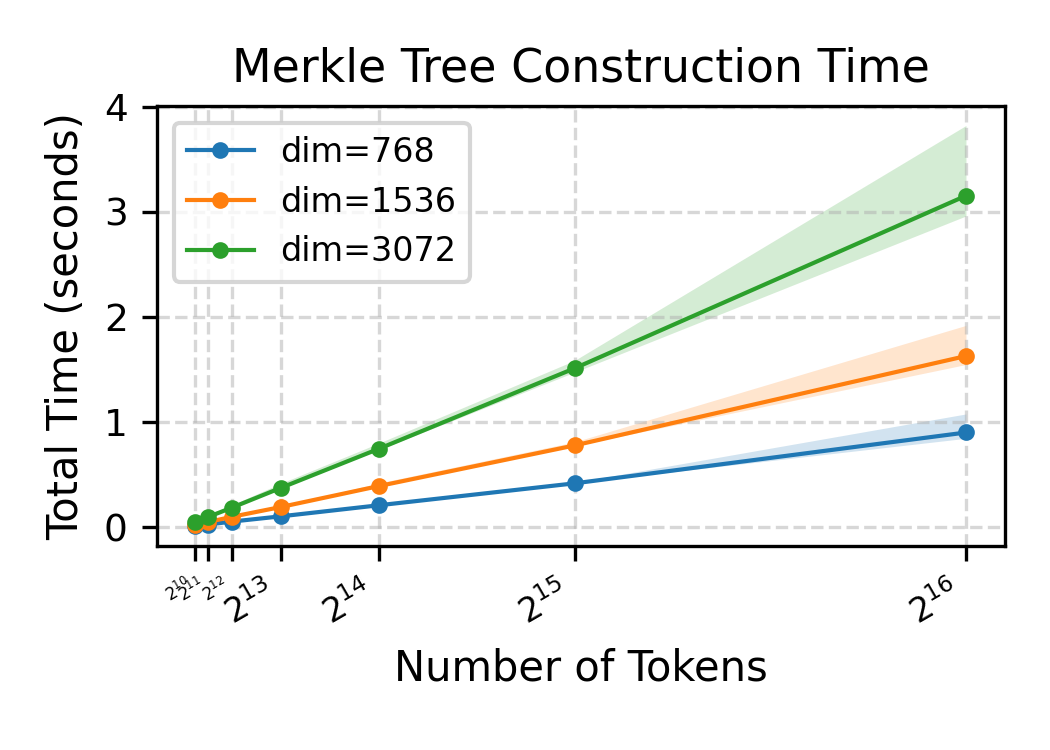}
        \caption{Merkle Tree Construction Time with Fluctuation Range.}
        \label{fig:merkle_time}
    \end{minipage}
    \hfill
    \begin{minipage}[c]{0.48\linewidth}
        \centering
        \includegraphics[width=\linewidth]{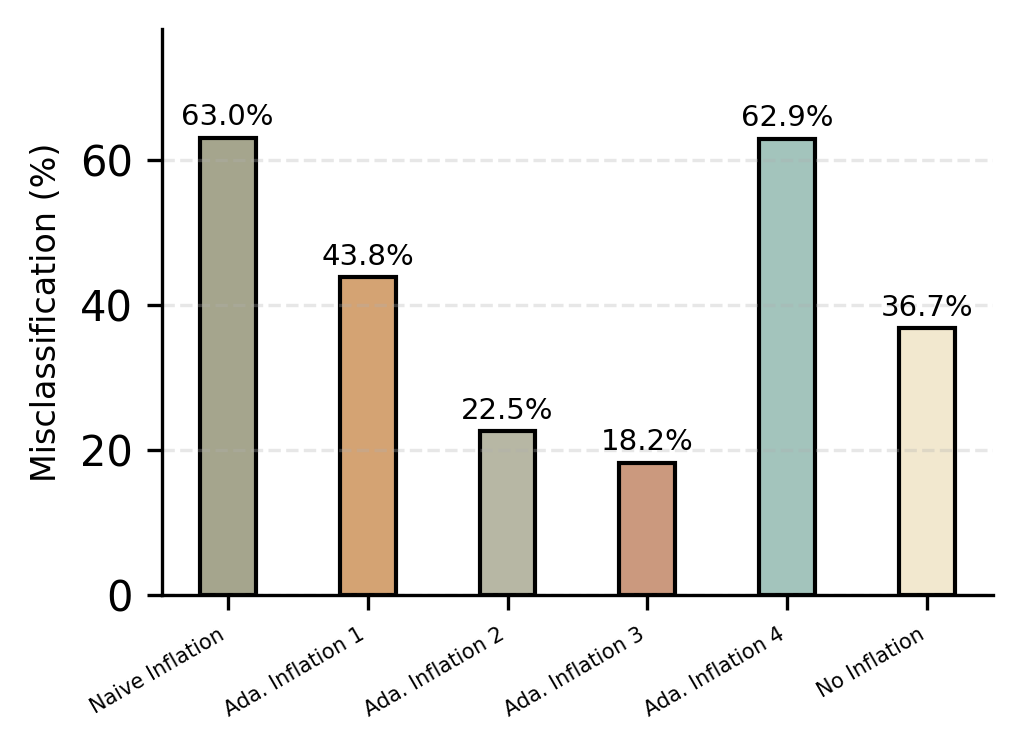}
        \caption{Misclassification Rates of LLMs on Constructed Datasets.}
        \label{fig:llm_misclass}
    \end{minipage}
\end{figure}

\noindent \textbf{Cost of Building Hash Trees.}
We evaluate the computational overhead of constructing the Merkle hash tree, with respect to input size and hidden dimension. 
Experiments were conducted on a dual-socket AMD EPYC 7763 system (128 cores, 256 threads). All constructions ran as single-threaded processes on one logical core.
As shown in Figure~\ref{fig:merkle_time}, the construction time grows approximately linearly with the input length for a fixed hidden dimension, and increases more steeply with higher dimensions. 
Given that most LLM inference servers have underutilized CPUs, and the Merkle Tree construction process scales effectively with multi-core parallelism, the practical cost of building the hash tree is nearly negligible.

\section{Discussion}

\noindent \textbf{Can the original text be recovered from the tokens and embeddings exposed by COLA?}
\begin{wraptable}{r}{0.45\linewidth}
    \centering
    \vspace{-10pt}
    \caption{Similarity Between Blocks Reconstructed by {\n} and Real Blocks.}
    \label{tab:reconstruct_block}
    \begin{tabular}{lccc}
        \toprule
        \multirow{2}{*}{Metric} & \multicolumn{3}{c}{Block Size $\beta$} \\
        \cmidrule(lr){2-4}
                                & 256 & 512 & 1024 \\
        \midrule
        EmbedSim         & 0.65 & 0.66 & 0.75 \\
        BLEU                  & 0.04 & 0.05 & 0.03 \\
        ROUGE-L               & 0.23 & 0.25 & 0.24 \\
        BERTScore         & 0.83 & 0.83 & 0.84  \\
        \bottomrule
    \end{tabular}
    \vspace{-10pt}
\end{wraptable}
During the verification process in {\n}, COLA leaks a certain number of block embeddings and tokens within the blocks to {\n}. 
To quantify the impact of such leakage, we assume a malicious {\n} leverages an RAG system to retrieve documents highly similar to the exposed embeddings and tokens, then feeds all retrieved information into an LLM to reconstruct the original content. The design and further details are provided in Appendix~\ref{app:prompt}. We randomly selected 100 samples from a mathematical dataset. We evaluated the similarity between the reconstructed blocks and the original ones using embedding similarity, BLEU score~\cite{papineni2002bleu}, ROUGE-L~\cite{lin2004rouge} , and BERTScore~\cite{zhang2019bertscore}. As shown in Table~\ref{tab:reconstruct_block}, we observe that a high BERTScore/EmbedSim combined with low BLEU/ROUGE indicates that the LLM successfully preserves the core semantics, while significantly differing from the real block in terms of surface expression and syntactic structure.

\noindent \textbf{How difficult is the dataset we constructed? }~\label{sec:dis-hard-data}
To investigate the dataset difficulty, we submitted the failed samples from the \textit{Block-to-Answer Verification} task, along with their Answer, to a LLM. 
Based on the idea of LLM-as-a-Judge~\cite{zheng2023judging,li2024llms}, we use a prompt to instruct the LLM to perform binary classification. The prompt used is provided in Appendix~\ref{app:prompt}.
The relatively high misclassification rate suggests that the LLM, after reading the original text, tends to align with the matching head's judgment. The LLM shows high error rates on Naive Inflation, Ada. Inflation 1 and 4, indicating strong performance of the matching head in these cases. However, it still struggles with the remaining two adaptive inflations. 
Notably, 36.7\% of real blocks were misclassified by the LLM, suggesting that some parts of the true reasoning steps may be unrelated to answer derivation.

\section{Limitations}

We acknowledge that {\n}, despite its merits, possesses certain limitations that warrant discussion.
\begin{itemize}
    \item Firstly, {\n} exhibits suboptimal performance in the detection of malicious samples when the Inflation Rate is low. However, it is pertinent to note that under such circumstances, the incentive for COLA to engage in data falsification is also correspondingly diminished.

    \item Secondly, {\n} is inherently probabilistic, and as such, it is susceptible to a non-zero misclassification rate. Consequently, when benign samples are erroneously identified as malicious, the protocol of {\n} necessitates that COLA discloses the original text to the user for verification.

    \item Thirdly, the auditing process facilitated by {\n} requires the active cooperation of COLA. Ideally, COLA itself could deploy {\n} to attest to its own integrity. This would allow COLA to continue concealing its reasoning tokens, thereby mitigating the risk of its proprietary model being subjected to distillation attacks.

    \item Finally, {\n} comprises multiple small-scale neural network components and is not architected as an end-to-end system. Nevertheless, this modular design confers a distinct advantage: it permits the independent training of each module, which significantly enhances both the convergence speed and the overall efficacy of the training process.
\end{itemize}

\section{Conclusion}

This paper presents {\n}, a novel auditing framework designed to verify the token counts and semantic validity of hidden reasoning traces in COLA. We identify and formalize the problem of \textit{token count inflation}, in which service providers can overcharge users by injecting redundant or fabricated reasoning tokens that are not visible to the user. To address this, {\n} integrates two complementary components: a hash tree-based token quantity verifier and a semantic relevance-based validity checker.
Our extensive experiments demonstrate that {\n} can detect both naive and adaptive inflation strategies with high accuracy, even under limited exposure settings. By enabling transparent and auditable billing without revealing proprietary content, {\n} introduces a practical mechanism for accountability in commercial LLM services. We hope this work lays the foundation for future research on LLM API auditing, transparent reasoning, and verifiable inference services.

\bibliography{reference}
\bibliographystyle{plainnat}


\appendix

\section{Dataset Construction and Experimental Details}\label{apdx:data}

\begin{algorithm}[H]
\caption{Streamlined Generation of Inflated Reasoning Sequences}
\label{alg:inflation_generation_appendix_condensed}
\begin{algorithmic}[1]
\Require Original dataset $D_{orig}$, inflation ratios $\mathcal{K}$, strategies $S_{list}$ with weights $W_S$, tokenizer $\mathcal{T}$, embedder $\mathcal{E}$, anchor source $Src_{anchor}$, segment length range $[L_{min}, L_{max}]$, insertion mode $M_{ins}$, and optional block range $[B_{min}, B_{max}]$ if using block mode.
\Ensure Inflated dataset $D_{inflated}$

\State Initialize $D_{inflated} \gets \emptyset$
\State Build FAISS indexes for RAG-based strategies

\For{each data point $item_i = (P_i, R_i, A_i)$ in $D_{orig}$}
    \State $T_{orig} \gets \mathcal{T}(R_i)$; \If{$T_{orig}$ is empty} \textbf{continue} \EndIf
    \State $T_{anchor} \gets \text{SelectAnchor}(item_i, Src_{anchor})$
    \State $N_{max} \gets \lfloor |T_{orig}| \cdot \max(\mathcal{K}) \rfloor$
    \State $T_{pool} \gets \text{CollectTokens}(N_{max}, T_{anchor}, S_{list}, W_S)$

    \For{each $k \in \mathcal{K}$}
        \State $N_k \gets \lfloor |T_{orig}| \cdot k \rfloor$
        \State $T_k \gets \text{Subsample}(T_{pool}, N_k)$
        \State $T_{final} \gets \text{Insert}(T_{orig}, T_k, M_{ins}, [B_{min}, B_{max}])$
        \State Add $\mathcal{T}^{-1}(T_{final})$ to $D_{inflated}$ with metadata
    \EndFor
\EndFor
\State \Return $D_{inflated}$
\end{algorithmic}
\end{algorithm}

\subsection{Dataset Construction Details}
We construct two verification datasets for Block-to-Answer and Token-to-Block verification, each dataset includes two types of inflated samples. The simple version consists entirely of artificially generated (inflated) tokens, while the hard version contains a mixture of real and inflated tokens. For Token-to-Block verification, we randomly sample between 3.125\% and 12.5\% of tokens from each block to create both training and test instances.

For both verification tasks, we generate 1,200,000 positive and negative samples respectively. The training set is uniformly distributed across four datasets. Since the difficulty levels of the samples vary, we adjust the composition using an adaptive inflation strategy (applied in Block-to-Answer) to ensure balanced learning.

For training the DeepSets model, we additionally sample 1,000 examples. To preserve generalization capability, the data used for training this model does not overlap with any samples seen by the matching heads.

\subsection{Experimental Details}
All evaluation results, unless stated otherwise, are reported on 1,000 examples. This applies to Block-to-Answer, Token-to-Block, and the test sets used within the {\n} framework. Each numeric result is computed over a minimum of 1,000 samples to ensure statistical significance. Please refer to the Algorithm~\ref{alg:inflation_generation_appendix_condensed} for our {\n} workflow test set construction process.

\subsection{Source of Dataset}
To evaluate {\n}'s performance across different domains, we constructed training and test sets based on five datasets distilled from \texttt{DeepSeek-R1}~\cite{deepseekai2025deepseekr1incentivizingreasoningcapability}, including Medical~\cite{chen2024huatuogpto1medicalcomplexreasoning}\footnote{\url{https://huggingface.co/datasets/FreedomIntelligence/Medical-R1-Distill-Data}}, Code~\cite{OpenThoughts,openr1}\footnote{\url{https://huggingface.co/datasets/open-r1/OpenThoughts-114k-Code_decontaminated }}, Math~\cite{openr1}\footnote{\url{https://huggingface.co/datasets/open-r1/OpenR1-Math-220k}}, General\footnote{\url{https://huggingface.co/datasets/glaiveai/reasoning-v1-20m }}, and Out-of-Domain data Math (OOD)~\cite{OpenThoughts,openr1}\footnote{\url{https://huggingface.co/datasets/open-r1/OpenThoughts-114k-math}}. Our final training set is a mixture of these five datasets.

\section{Training and Model Details}\label{apdx:train}

For the matching heads used in Token-to-Block verification and Block-to-Answer verification, we set the learning rate to $2 \times 10^{-5}$, the batch size to 128, and train for 3 epochs. We employ the Adam optimizer and use the Focal Loss function. The hidden dimension of the model follows that of the embedding model, set to 384.

For the DeepSets model in the verifier, we use a batch size of 128, a hidden dimension of 256, and train for 5 epochs. We adopt the Adam optimizer with a learning rate of $1 \times 10^{-3}$ and use the binary cross-entropy (BCE) loss. All experiments are conducted with a fixed random seed of 42.

\section{Computational Resources}

All experiments were conducted on a high-performance workstation running Ubuntu 20.04.6 LTS. The system is equipped with a dual-socket AMD EPYC 7763 processor, providing a total of 128 physical cores and 256 threads. For GPU acceleration, we utilized an NVIDIA RTX A6000 Ada graphics card.

\section{Details of {\n}}

\subsection{Merkle Tree Construction}\label{app:tree}

Algorithm~\ref{alg:merkle_construction_cola} details the process COLA uses to construct the Merkle Hash Tree from a reasoning sequence $R$. This corresponds to the "Token Fingerprint Generation" and "Fingerprint Hash Tree Construction" paragraphs.

\subsection{Merkle Proof Verification}\label{app:proof}

Algorithm~\ref{alg:merkle_proof_cola} describes how the {\n} auditor verifies the integrity of a token using its fingerprint and the Merkle path provided by COLA. This corresponds to the "Merkle Proof" paragraph.

\begin{algorithm}[H]
\caption{Merkle Proof Verification}
\label{alg:merkle_proof_cola}
\begin{algorithmic}[1]
\Require Committed Merkle Root $MR_{committed}$ (from COLA).
\Require Token fingerprint $fp_{token}$ of the audited token (from COLA).
\Require Merkle Path $P = [(h_1, pos_1), (h_2, pos_2), \ldots, (h_d, pos_d)]$ (from COLA), where $h_k$ is a sibling hash and $pos_k \in \{\text{`left'}, \text{`right'}\}$ indicates $h_k$'s position relative to the path node.
\Require Cryptographic hash function $H(\cdot)$.
\Ensure Boolean: \textbf{true} if verification succeeds, \textbf{false} otherwise.

\State $current\_computed\_hash \gets H(fp_{token})$ \Comment{Hash the provided token fingerprint}

\For{each pair $(sibling\_hash, position) \in P$}
    \If{$position = \text{`left'}$}
        \State $current\_computed\_hash \gets H(sibling\_hash \concat current\_computed\_hash)$
    \ElsIf{$position = \text{`right'}$}
        \State $current\_computed\_hash \gets H(current\_computed\_hash \concat sibling\_hash)$
    \Else
        \State \Return \textbf{false} \Comment{Error: Invalid position in Merkle Path}
    \EndIf
\EndFor

\State $MR_{recomputed} \gets current\_computed\_hash$

\If{$MR_{recomputed} = MR_{committed}$}
    \State \Return \textbf{true} \Comment{Verification successful: token integrity confirmed}
\Else
    \State \Return \textbf{false} \Comment{Verification failed: mismatch indicates potential issue}
\EndIf
\end{algorithmic}
\end{algorithm}

\paragraph{Notes on Algorithms:}
\begin{itemize}
    \item \textbf{Padding (Algorithm~\ref{alg:merkle_construction_cola}):} The text states, "The number of leaf nodes is padded to the nearest power of two." Algorithm~\ref{alg:merkle_construction_cola} implements this by duplicating the hash of the last actual leaf node if leaves exist. If the initial set of tokens (and thus fingerprints) is empty ($N=0$), it assumes padding to $N_{pow2}=1$ using a hash of a predefined value (e.g., an empty string). The exact nature of this padding for an empty set should be consistently defined between COLA and the auditor.
    \item \textbf{Merkle Path Representation (Algorithm~\ref{alg:merkle_proof_cola}):} The Merkle Path $P$ is assumed to be a list of (hash, position) tuples. The `position` indicates if the sibling hash is to the 'left' or 'right' of the node on the direct path from the audited leaf to the root.
    \item \textbf{Concatenation for Hashing:} The order of concatenation (e.g., $H(leftChild \concat rightChild)$ vs. $H(rightChild \concat leftChild)$) must be consistent throughout construction and verification. The algorithms assume a fixed order (left child first).

\end{itemize}

\begin{algorithm}[H] 
\caption{Merkle Tree Construction by COLA}
\label{alg:merkle_construction_cola}
\begin{algorithmic}[1] 
\Require Reasoning tokens $R$; number of blocks $\alpha$; embedding function $\Embd(\cdot)$; cryptographic hash function $H(\cdot)$.
\Ensure Merkle Root $MR$.

\SectionComment{Phase 1: Token Fingerprint Generation and Leaf Node Creation}
\State $Blocks \gets \text{Partition}(R, \alpha)$ \Comment{Partition $R$ into $B_1, \ldots, B_\alpha$}
\State $Fingerprints \gets \emptyset$ \Comment{Initialize as an empty list}
\For{each block $B_j \in Blocks$}
    \State $e_{block_j} \gets \Embd(B_j)$ \Comment{Compute block embedding}
    \For{each token $r_i \in B_j$}
        \State $e_{token_i} \gets \Embd(r_i)$ \Comment{Compute token embedding}
        \State $fp_i \gets e_{block_j} \concat e_{token_i}$ \Comment{Form token fingerprint}
        \State Add $fp_i$ to $Fingerprints$
    \EndFor
\EndFor

\State $LeafNodes \gets \emptyset$ \Comment{Initialize as an empty list}
\For{each fingerprint $fp \in Fingerprints$}
    \State $leaf \gets H(fp)$ \Comment{Hash fingerprint to create leaf node}
    \State Add $leaf$ to $LeafNodes$
\EndFor

\SectionComment{Phase 2: Padding Leaf Nodes}
\State $N \gets \text{length}(LeafNodes)$
\State Let $N_{pow2}$ be the smallest power of two such that $N_{pow2} \ge N$.
\If{$N < N_{pow2}$}
    \If{$N = 0$ \textbf{and} $N_{pow2} > 0$} \Comment{e.g., $N=0 \implies N_{pow2}=1$}
        \State $padding\_hash \gets H(\text{""})$ \Comment{Hash of empty string or other predefined padding value}
        \For{$k \gets 1 \textbf{ to } N_{pow2}$}
            \State Add $padding\_hash$ to $LeafNodes$
        \EndFor
    \ElsIf{$N > 0$}
        \State $last\_leaf\_hash \gets LeafNodes[N-1]$ \Comment{Get hash of the last actual leaf}
        \For{$k \gets 1 \textbf{ to } N_{pow2} - N$}
            \State Add $last\_leaf\_hash$ to $LeafNodes$ \Comment{Pad by duplicating the last leaf's hash}
        \EndFor
    \EndIf
\EndIf

\SectionComment{Phase 3: Building the Tree Recursively}
\State $CurrentLevelNodes \gets LeafNodes$
\While{$\text{length}(CurrentLevelNodes) > 1$}
    \State $NextLevelNodes \gets \emptyset$
    \For{$k \gets 0 \textbf{ to } (\text{length}(CurrentLevelNodes)/2) - 1$}
        \State $leftChild \gets CurrentLevelNodes[2k]$
        \State $rightChild \gets CurrentLevelNodes[2k+1]$
        \State $parentHash \gets H(leftChild \concat rightChild)$
        \State Add $parentHash$ to $NextLevelNodes$
    \EndFor
    \State $CurrentLevelNodes \gets NextLevelNodes$
\EndWhile

\If{$\text{length}(CurrentLevelNodes) = 1$}
    \State $MR \gets CurrentLevelNodes[0]$ \Comment{The single remaining node is the Merkle Root}
\Else \Comment{Handles $N=0$ and $N_{pow2}=0$, resulting in an empty $CurrentLevelNodes$}
    \State $MR \gets H(\text{""})$ \Comment{Define Merkle Root for an empty set of tokens, e.g., hash of empty string}
\EndIf
\State \Return $MR$
\end{algorithmic}
\end{algorithm}

\subsection{Workflow of {\n}}

Algorithm~\ref{alg:multi_round_verification} illustrates the complete verification procedure of {\n}.

\begin{algorithm}[H] 
\footnotesize 
\caption{Multi-Round Verification in {\n}}
\label{alg:multi_round_verification}
\begin{algorithmic}[1] 
\Require COLA Response (containing reasoning blocks $\mathcal{B}_{\text{total}}$ and final answer $A$)
\Require Fraction $\gamma$ of blocks for initial verification (e.g., 0.1)
\Require Pre-trained matching heads $\mathsf{MH_{tb}}(\cdot, \cdot)$, $\mathsf{MH_{ba}}(\cdot, \cdot)$
\Require Embedding function $\mathsf{Embd}(\cdot)$
\Require Verification threshold $\tau$
\Ensure Audit decision: "Successful" or "COLA Flagged for Inflation"

\Statex \textit{// Initialization}
\State $\mathcal{B}_{\text{unverified}} \gets \mathcal{B}_{\text{total}}$
\State $\mathcal{B}_{\text{verified}} \gets \emptyset$
\State $\textit{audit\_successful} \gets \textbf{false}$
\State $\textit{all\_blocks\_audited} \gets \textbf{false}$

\Statex \textit{// Initial round of verification}
\State Select an initial set of blocks $\mathcal{B}_{\text{current\_round}} \subseteq \mathcal{B}_{\text{unverified}}$ of size $\lceil \gamma \cdot |\mathcal{B}_{\text{total}}| \rceil$
\If{$\mathcal{B}_{\text{current\_round}}$ is empty \textbf{and} $|\mathcal{B}_{\text{total}}| > 0$}
    \State $\mathcal{B}_{\text{current\_round}} \gets \text{one randomly selected block from } \mathcal{B}_{\text{unverified}}$
\EndIf

\While{\textbf{not} $\textit{audit\_successful}$ \textbf{and not} $\textit{all\_blocks\_audited}$}
    \If{$\mathcal{B}_{\text{current\_round}}$ is empty}
        \State $\textit{all\_blocks\_audited} \gets \textbf{true}$ \Comment{No more blocks to check}
        \State \textbf{goto} \textit{FinalDecision} 
    \EndIf

    \State $\textit{round\_scores} \gets []$ \Comment{Initialize as an empty list/array}
    \For{each block $B_j \in \mathcal{B}_{\text{current\_round}}$}
        \State Randomly select a subset of reasoning tokens $\{r_i\}_{i=1}^k$ from $B_j$ (e.g., 10
        \State $E_{\text{tokens}} \gets \mathsf{AVG}(\{\mathsf{Embd}(r_i)\}_{i=1}^k)$ \Comment{Average embedding of selected tokens}
        \State $E_{\text{block}} \gets \mathsf{Embd}(B_j)$
        \State $E_{\text{answer}} \gets \mathsf{Embd}(A)$

        \State $S_{tb} \gets \mathsf{MH_{tb}}(E_{\text{tokens}}, E_{\text{block}})$ \Comment{Token-to-Block score}
        \State $S_{ba} \gets \mathsf{MH_{ba}}(E_{\text{block}}, E_{\text{answer}})$ \Comment{Block-to-Answer score}
        \State Add pair $(S_{tb}, S_{ba})$ to $\textit{round\_scores}$
        \State {\n}\ performs Merkle Proofs on selected tokens in $B_j$ (verification of token integrity)
    \EndFor

    \State $\mathcal{B}_{\text{verified}} \gets \mathcal{B}_{\text{verified}} \cup \mathcal{B}_{\text{current\_round}}$
    \State $\mathcal{B}_{\text{unverified}} \gets \mathcal{B}_{\text{unverified}} \setminus \mathcal{B}_{\text{current\_round}}$

    \State $\textit{verifier\_decision} \gets \Call{Verifier}{\textit{round\_scores}, \tau}$
    \Comment{Verifier can be rule-based or learning-based}
    \If{$\textit{verifier\_decision} = \text{Accept}$}
        \State $\textit{audit\_successful} \gets \textbf{true}$
    \Else
        \If{$\mathcal{B}_{\text{unverified}}$ is empty}
            \State $\textit{all\_blocks\_audited} \gets \textbf{true}$
        \Else
            \State Select one new random block $B_{\text{next}}$ from $\mathcal{B}_{\text{unverified}}$
            \State $\mathcal{B}_{\text{current\_round}} \gets \{B_{\text{next}}\}$ \Comment{Next round verifies this single block}
        \EndIf
    \EndIf
\EndWhile

\Statex \textit{FinalDecision:} \label{FinalDecision} 
\If{$\textit{audit\_successful}$}
    \State \Return "Audit Successful"
\Else
    \State \Return "COLA Flagged for Token Inflation"
    \Comment{User may request COLA to justify charges}
\EndIf

\Statex
\Function{Verifier}{$\textit{scores\_list}, \tau$}
    \Comment{Example: Rule-based verifier}
    \If{$\textit{scores\_list}$ is empty} \Return "Reject" \EndIf
    \State $\textit{avg\_S\_tb} \gets \text{average of all } S_{tb} \text{ in } \textit{scores\_list}$
    \State $\textit{avg\_S\_ba} \gets \text{average of all } S_{ba} \text{ in } \textit{scores\_list}$
    \If{$\textit{avg\_S\_tb} > \tau$ \textbf{and} $\textit{avg\_S\_ba} > \tau$}
        \State \Return "Accept"
    \Else
        \State \Return "Reject"
    \EndIf
    \Comment{Alternatively, a learning-based verifier (e.g., DeepSets) could be used here.}
\EndFunction
\end{algorithmic}
\end{algorithm}

\section{Detection Performance of {\n}}\label{apdx:blocks}

\begin{figure}[tbp]
    \centering
    \includegraphics[width=1.07\textwidth]{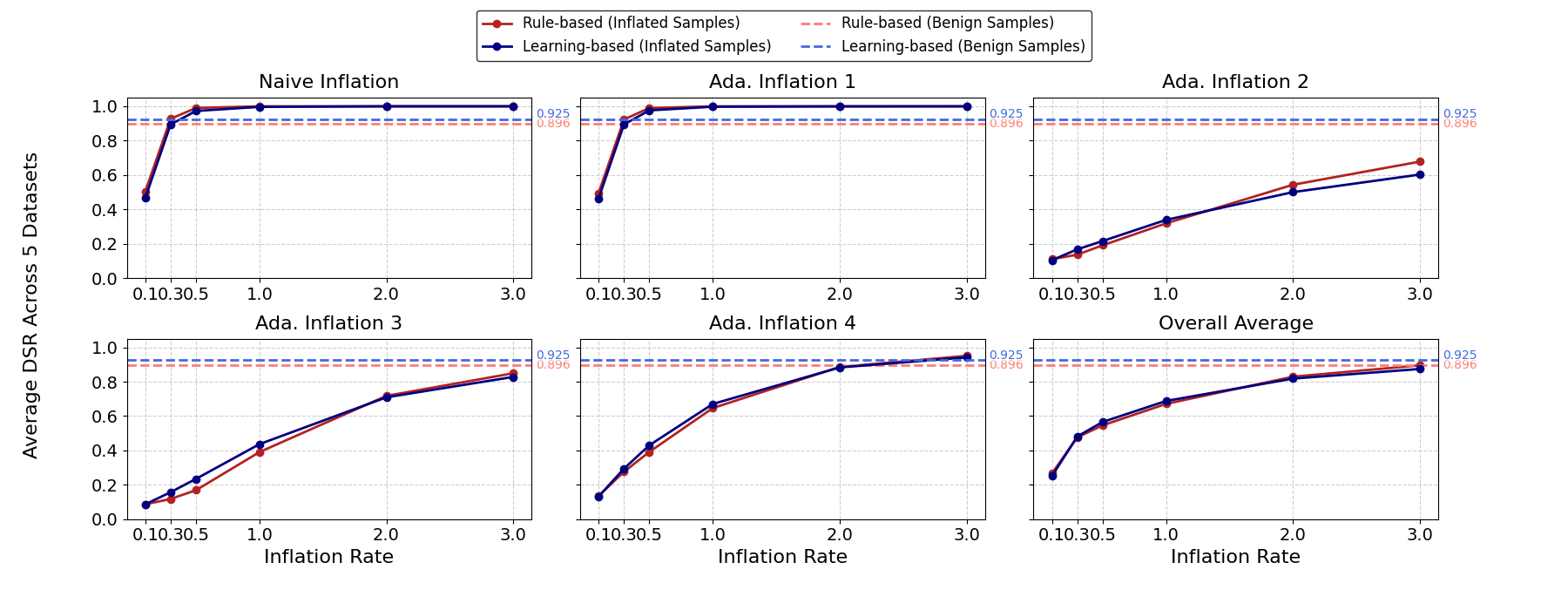} 
    \caption{Performance of {\n} across different inflation methods and verifiers (Block Size = 512). The red lines and the blue lines represent the DSR of rule-based verifier and learning-based verifier, respectively. $\gamma$
    }
    \label{fig:main_coin1_512}
\end{figure}

\begin{figure}[tbp]
    \centering
    \includegraphics[width=1.05\textwidth]{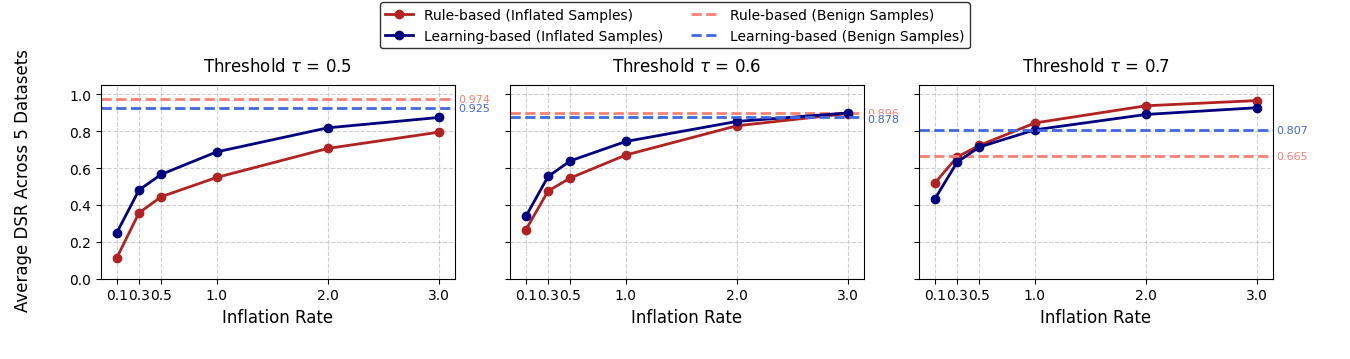} 
    \caption{Impact of threshold $\tau$ on DSR (Block Size = 512). 
    }
    \label{fig:main_coin2_512}
\end{figure}

\begin{figure}[tbp]
    \centering
    \includegraphics[width=1.07\textwidth]{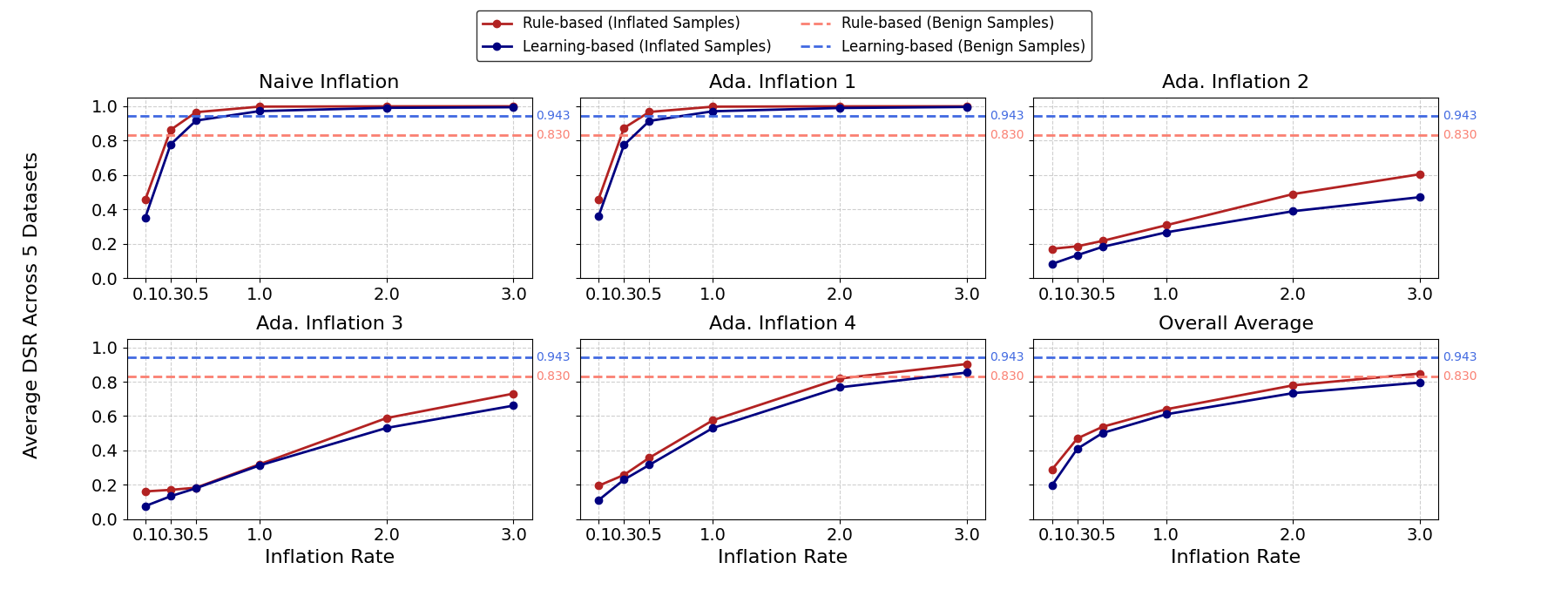} 
    \caption{Performance of {\n} acorss different inflation methods and verifiers (Block Size = 1024). The red lines and the blue lines represent the DSR of rule-based verifier and learning-based verifier, respectively. $\gamma$
    }
    \label{fig:main_coin1_1024}
\end{figure}

\begin{figure}[tbp]
    \centering
    \includegraphics[width=1.05\textwidth]{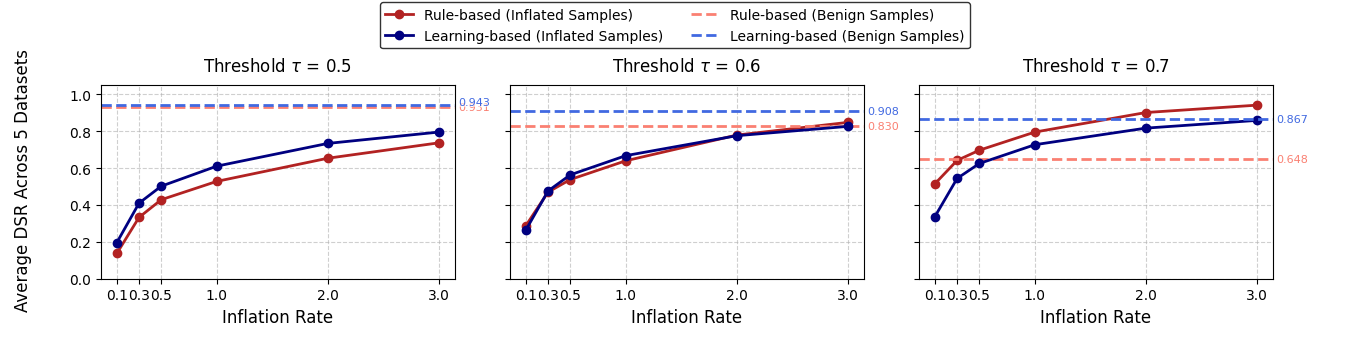} 
    \caption{Impact of threshold $\tau$ on DSR (Block Size = 1024). 
    }
    \label{fig:main_coin2_1024}
\end{figure}

We show the comparison of the two verifiers and the impact of $\tau$ under different block sizes, as shown in Figure~\ref{fig:main_coin1_512},\ref{fig:main_coin2_512},\ref{fig:main_coin1_1024},\ref{fig:main_coin2_1024}

\section{Prompts Used in Discussion Section}\label{app:prompt}

Prompt~\ref{fig:reasoning_recovery_prompt} is used to explore the question ``Can the original text be recovered from the tokens and embeddings exposed by COLA?'', while Prompt~\ref{fig:reasoning_evaluation_prompt} is used to explore ``How difficult is the dataset we constructed?''.

\begin{figure}[!ht]
\begin{prompt}{Prompt for Reconstructing Hidden Reasoning Passage}
\textbf{Reconstructing Hidden Reasoning Prompt}\\
You are an expert in natural language reasoning and semantic retrieval.\\
Your task is to help recover a semantically meaningful and logically connected hidden passage that bridges a <Question> and an <Answer>.\\
This passage has been lost, but we know it is semantically related to both the <Question> and the <Answer>, and lies between them.\\

Given a -- \verb|**|<Question>\verb|**|:\\
\{question\}\\

And the -- \verb|**|<Answer>\verb|**|:\\
\{answer\}\\

We also know that some tokens from the original passage are still visible:\\
\{sampled\_token\_text\}\\

And we retrieved related documents from Wikipedia using the embedding of the original passage:\\
\{retrieved\_rag\_docs\}\\

Now, please help recover the most likely content of the hidden passage.\\
Return your answer strictly in the following JSON format:\\

\texttt{
\textbackslash\textbackslash recovered\_json
\{\\
\hspace*{4mm}"recovered\_text": "<your reconstructed passage here>"\\
\}
}
\end{prompt}
\caption{Prompt for Recovering a Hidden Reasoning Passage Using Question, Answer, Token Clues and Retrieved Wikipedia Documents.}
\label{fig:reasoning_recovery_prompt}
\end{figure}

\begin{figure}[!ht]
\begin{prompt}{Prompt for Evaluating Reasoning Passage Relevance}
\textbf{Evaluating Reasoning Process Prompt}\\
You are a logical reasoning analyst.\\

Given a final answer and a randomly selected text passage, your task is to assess whether the text passage represents a reasoning process that leads to or supports the final answer.\\

The passage may or may not be relevant to the answer.\\
Your task is \textbf{not} to verify factual correctness, but to determine whether the passage semantically or logically connects to the answer and explains or justifies it in any meaningful way.\\

\verb|**|Random Text Passage\verb|**|:\\
\{reason\}\\

\verb|**|Final Answer\verb|**|:\\
\{answer\}\\

Please answer the following questions:\\
1. Is the text passage a plausible reasoning process that leads to the final answer?\\
2. Does it provide logical or semantic justification for the answer?\\

Respond in the following JSON format:\\

\texttt{
\textbackslash\textbackslash reasoning\_assessment\\
\{\\
\hspace*{4mm}"is\_reasoning\_process": true/false,\\
\hspace*{4mm}"justification": "<your brief explanation of why the passage is or isn’t a reasoning process for the answer>"\\
\}
}
\end{prompt}
\caption{Prompt for Judging Whether a Block Supports or Explains a Final Answer.}
\label{fig:reasoning_evaluation_prompt}
\end{figure}

\end{document}